\documentclass[runningheads]{llncs}
\usepackage[T1]{fontenc}
\usepackage{graphicx,verbatim}
\usepackage{subcaption} 
\usepackage{amsmath}

\usepackage{multirow}
\usepackage{array}
\usepackage{tabularx}

\newcolumntype{C}{>{\centering\arraybackslash}X}
\begin{document}
\title{Morphology signal in whole slide image foundation models can automatically triage slides}
\titlerunning{Morphology signal in WSI FMs can automatically triage slides}

\author{Ayushi Sinha\inst{1} \and 
Shashank Yadav\inst{2} \and 
Benjamin Holmes\inst{3} \and
Pravat Das\inst{3} \and
Aaron W. Bogan\inst{4} \and 
James S. Lewis Jr.\inst{5} \and 
Santiago Romero-Brufau\inst{6} \and 
Andrew Y. K. Foong\inst{2} \and 
Scott H. Kaufmann\inst{7} \and 
Kathryn M. Van Abel\inst{6} \and 
David M. Routman\inst{2} \and 
Michael R. Lucas\inst{3} }

\authorrunning{A. Sinha et al.}
%

\institute{Department of Radiology, Mayo Clinic, Rochester MN \and 
Department of Radiation Oncology, Mayo Clinic, Rochester MN \and 
AI Program, Mayo Clinic, Rochester MN \and 
Department of Quantitative Health Sciences, Mayo Clinic, Scottsdale AZ \and 
Department of Laboratory Medicine and Pathology, Mayo Clinic, Scottsdale AZ \and 
Department of Otolaryngology-Head and Neck Surgery, Mayo Clinic, Rochester MN \and
Department of Oncology, Mayo Clinic, Rochester MN \\
\email{lucas.michael2@mayo.edu}}

\maketitle              
\begin{abstract}
Patient exams in the cancer diagnosis and staging process typically generate several whole slide images (WSIs). One of the initial steps in training models on WSI data is identifying one or a few slides containing tumor or other diagnostic biomarkers necessary for downstream prediction tasks such as estimating recurrence risk or progression-free survival. This step requires tedious manual curation by experienced pathologists. Many published datasets make the artificial assumption of $1$ slide per patient. Alternatively, all slides per patient may be used for model training, which may dilute the signal from the few slides containing tumor or other relevant information. In this paper, we present a pipeline to overcome these challenges using publicly available WSI foundation models (FMs). Our evaluations show that ranking WSIs based on predictions from zero-shot classification using WSI FMs accurately identifies slides with the most tumor, indicating that WSI FMs contain sufficient morphology signal to automatically triage slides. We also present a formulation for ranked evaluation to benchmark FM performance in slide triage. We show, on multiple datasets, that tumor slides are identified in the top-$2$ ranked slides for patients with up to $43$ slides. 


\end{abstract}
\section{Introduction}
Cancer diagnosis and staging typically involve biopsy and resection specimens, which are prosected, sectioned, processed into paraffin-embedded blocks, processed into glass slide sections, and finally reviewed by the pathologist for diagnosis.  With the growth of digital pathology, slides are increasingly being digitized to generate whole slide images (WSIs). This process typically generates a large number of WSIs per patient, particularly for resection specimens. However, only a few WSIs contain tumor or other diagnostic biomarkers that may be useful for downstream prediction tasks such as estimating recurrence risk or progression-free survival. Current practice requires a pathologist to triage all generated WSIs per patient by visually analyzing and identifying those with primary tumor and/or lymph nodes involved by metastasis, which can take anywhere from about $30-90s$ per slide~\cite{hanna_whole_2019}. Automatically identifying the one or few most relevant slides per patient is a crucial, albeit overlooked, step in WSI model training. In this paper, we show that foundation models (FMs) trained jointly on histopathology image and text data may contain sufficient morphology signal to automatically triage slides without the need for manual curation.

\begin{figure}[t]
\centering
\includegraphics[width=\textwidth]{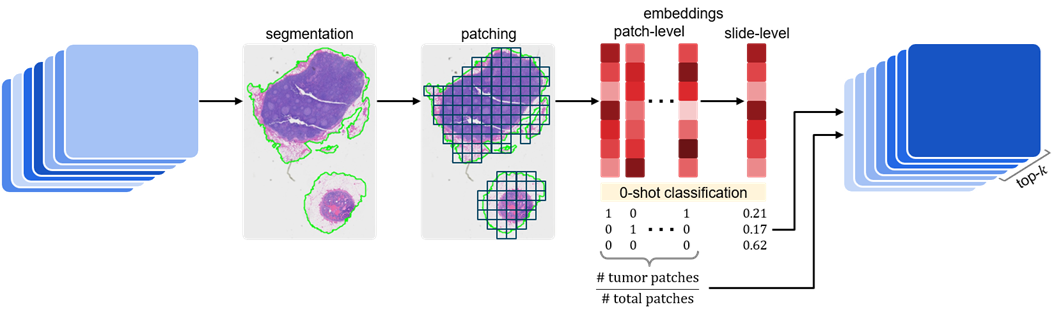}
\caption{Schematic diagram of the presented pipeline which takes takes all slides per patient and ranks them to enable selection of the top-$k$ slides. The shade of blue depicts, as an example, the percent tumor in each slide with darker blue denoting more tumor. From left to right, the slides are first segmented to remove background pixels, after which patches are extracted from the tissue regions. FMs are used to generate image embeddings at the patch- and slide-level, which are then used along with text embeddings generated from class specific prompts to classify the image. At the patch-level, classification results are aggregated by computing the percent patches in a class per slide, which are then used to rank the slides. At the patch-level, class probabilities are used directly to rank slides.} \label{cover}
\end{figure}

WSIs present several computational challenges. A typical WSI is digitized at various magnifications ranging from $0.25$ microns per pixel (mpp) to $2$mpp. At the highest magnification levels, images may be tens of thousands of pixels in resolution. To process images of such magnitude, WSIs are split into smaller patches (e.g., $256 \times 256$~px or $512 \times 512$~px). Histopathology models are subsequently trained at the patch level, with FMs generating patch-level embeddings~\cite{chen2024uni,lu2024avisionlanguage,vorontsov2024virchow,alber_atlas_2025,alber_atlas_2026}. Patch-level embeddings can be aggregated across WSIs to generate slide-level embeddings~\cite{ding2025multimodal,shaikovski2024prism,xu2024gigapath,wang_pathology_2024,zhang2026care}. Downstream training tasks, such as predicting recurrence risk, also typically use patch-level embeddings along with aggregation methods such as multiple instance learning (MIL) to learn slide-level labels~\cite{dietterich_solving_1997}. Attention-based MIL (AB-MIL) proposes using attention to create weighted representations for each slide or patient by learning which patches contribute most to the prediction~\cite{pmlr-v80-ilse18a}. When datasets have not been manually curated to select a single slide per patient for prediction tasks, patches from all slides may be used~\cite{yao2020whole,shao2023hvtsurv}. However, each slide can generate thousands of patches, diluting the signal from the few patches containing tumor or other relevant information.

To overcome this limitation, we perform zero-shot classification using publicly available pathology FMs by computing cosine similarity between image and text embeddings to assess tumor content in WSIs. While FMs have been used to classify slides both by fine-tuning~\cite{guo2026clinically} and using zero-shot classification~\cite{ahmed2024pathalign}, these have primarily been done for cancer subtyping and not for slide triage. Using the pipeline presented in this paper (Fig.~\ref{cover}), we explore whether pathology FMs contain sufficient morphology signal to automatically triage slides. We evaluate patch- and slide-level embeddings, and present a formulation for ranked evaluation that takes into consideration the number of slides per patient, the number of slides with tumor content, and the rank of the accurately identified slides. This formulation is used to benchmark models for slide triage on $2$ datasets. We show that by ranking WSIs by tumor content, we can accurately identify slides with the most tumor, demonstrating that automatic slide triage is possible.

\section{Methods}
\subsection{Datasets}
Two datasets consisting of WSIs and accompanying pathology reports were used to evaluate our pipeline. Pathology reports were used to extract annotations that were used for evaluation. Pathology reports are organized by anatomical specimens received into \emph{parts}, which are further organized, after sectioning, processing, and embedding, into \emph{blocks} of tissue. Parts are assigned letters and blocks are assigned numerical values to generate block codes (e.g., A1, A2, B5, etc.). Each block may generate multiple slides, although most blocks in our datasets only contained a single slide. Each slide is identified only by its block code. A single pathology report representing findings from all slides is generated per patient exam. The presence or absence of cancer is typically noted at the part level. Details of the two datasets are provided below:

\textbf{HN}: The head \& neck cancer dataset was compiled from $1233$ patients consisting of $55597$ WSIs. These slides were drawn from patients with cancer specimens mainly from the neck, tonsils, tongue, pharynx, and cervical lymph nodes. Slide-level annotations (e.g., slide with most tumor, etc.) were not available for this dataset. However, pathology reports were used to extract coarse part-level annotations. That is, if part A was positive for cancer, then all blocks in part A (i.e., A1, A2, \ldots) were labeled as cancer positive, although only one or few of these blocks may actually be cancer positive. When pathology reports identified cancer positivity at specific complexity levels, the annotation was refined to only include block codes associated with those complexity levels within the part. 

For a subset of the data, lymph nodes involved by metastases were identified at the block-level in the pathology report. This enabled us to curate a \textbf{HN-node} subset of the HN dataset consisting of $3960$ lymph node slides where, for $199$ patients, involved lymph nodes were identified at a more granular level than in the full HN dataset.

 \textbf{OV}: The ovarian cancer dataset was compiled from $35$ patients consisting of $1068$ slides drawn mainly from the uterus, ovaries, peritoneum, omentum, and pelvic lymph nodes. It was manually curated by a physician by visually evaluating each slide and identifying one or few blocks per patient with the most tumor content. The block with the most tumor content was annotated as the primary block, while any additional blocks identified per patient as also containing tumor were annotated as secondary blocks. All annotations, verified by a pathologist, were used only for evaluation.

\subsection{Data processing and embedding generation}
All WSIs were segmented using the HEST segmenter~\cite{jaume2024hest} to remove background pixels, and $512 \times 512$ patches were extracted from the tissue region at $20\times$ magnification ($0.5$mpp). Embeddings were generated using the following FMs:

\begin{itemize}
\item \textbf{CONCH}, which uses contrastive learning objectives in a ViT-based architecture to align image and text modalities into a shared embedding space~\cite{lu2024avisionlanguage}. CONCH generates patch-level embeddings.

\item \textbf{mean-CONCH}, which mean-pools patch-level CONCH embeddings into slide-level embeddings.

\item \textbf{TITAN}, which leverages patch-level embeddings from CONCH v1.5 in a vision-only self-supervised learning paradigm and fine-tunes the resulting model using contrastive alignment with text data~\cite{ding2025multimodal}. TITAN generates slide-level embeddings.
\end{itemize}

\subsection{Zero-shot classification and slide ranking}
Zero-shot classification was performed using text prompts designed to classify image embeddings into the following classes: \{\texttt{tumor, lymphocyte, other}\}. Prompts for \texttt{tumor} and \texttt{other} classes were adapted per cancer type (Tables~\ref{tumor_prompts}, \ref{lymphocyte_prompts},~\ref{other_prompts}). An ensemble of prompts was used for each class to account for the variability in verbiage across pathologists (Table~\ref{prompt_templates}). Text embeddings were generated for each prompt and a mean embedding was computed per class.

Cosine similarity was computed between image and text embeddings and class probabilities were calculated by applying \texttt{softmax} over the similarity scores. Class predictions were assigned to the class with the highest probability. For mean-CONCH and TITAN, class probabilities were used directly to rank slides. Patch-level embeddings from CONCH were further aggregated using class predictions per patch to compute the percent patches per slide assigned to a class, which was then used to rank slides. The top-$k$ slides were selected based on these rankings and any slides with class probability $0$ or $0\%$ patches in a class are excluded from the top-$k$. Therefore, a top-$k$ selection can generate a set of slides smaller than $k$.

\section{Evaluation}
\subsection{Recall@$k$}
We use recall@$k$ to evaluate whether any of the annotated block codes are included in the top-$k$ slides ranked by the $3$ models. We use a variant of recall@$k$ common in retrieval tasks, where recall per patient is defined as:

\begin{equation}
    \texttt{recall}^k_i = 
    \begin{cases}
        1 & \text{if at least 1 annotated block code is in the top-}k \\
        0 & \text{otherwise}
    \end{cases}
\end{equation}
Recall@$k$ across a dataset of $N$ patients can then be calculated as:
    \begin{equation}
        \texttt{recall}^k = \frac{\sum_{0<i<N} \texttt{recall}^k_i}{N}
    \end{equation}
This evaluation generates insights into the number of patients in a dataset for whom an annotated block code was in the top-$k$ ranked slides.

\subsection{Agreement between models}
Agreement among models is computed at each $k$ to assess whether they produce the same or similar ranking across patients. Agreement is calculated by computing the intersection of the sets of top block codes ranked by each model at each $ k$, and taking the cardinality of the intersection. This value is normalized by the cardinality of the ideal intersected set, which is $k$ $\times$ \texttt{num\_patients} for each $k$.

\subsection{Ranked evaluation}
\begin{figure}[t]
\includegraphics[width=\textwidth]{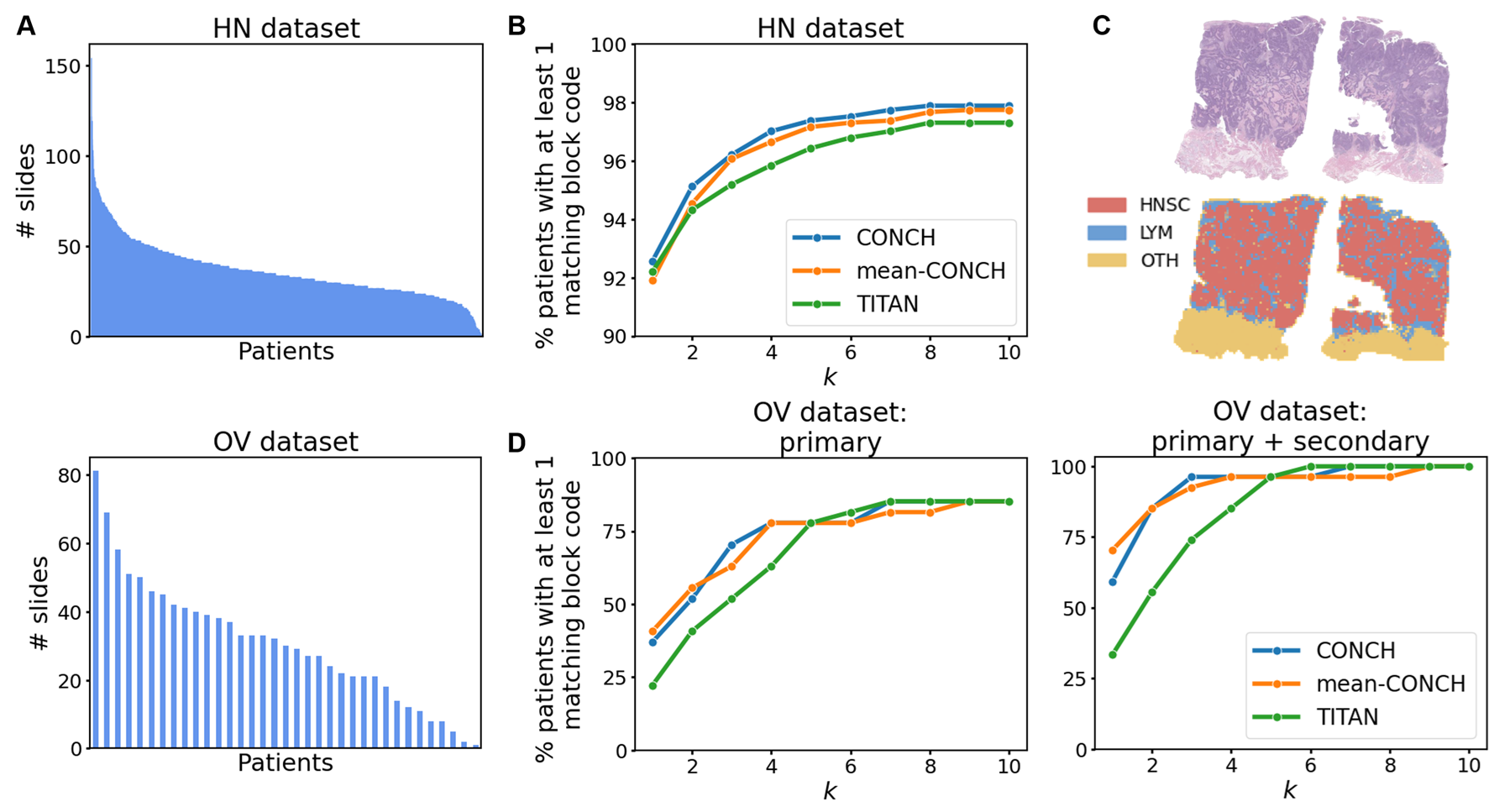}
\caption{A: Distribution of slides per patient across the HN (top) and OV (bottom) datasets. The HN dataset has slides in the range $[1,154]$ per patient with an average of $36.4\pm15.8$ slides per patient, while the HN-node dataset has slides in the range $[6,54]$ per patient with an average of $19.9\pm8.9$ slides per patient. The OV dataset has slides in the range $[1,81]$ per patient with an average of $30.5\pm18.4$ slides per patient. B: All models show similarly high recall at all $k$ on the HN dataset, where slides with tumor are labeled coarsely at the part level. C: Patch-level classification from CONCH aligns accurately with slide morphology. D: On the OV dataset, where only $1$ or a few slides with tumor are annotated per patient, recall at lower $k$ is low but quickly increases with higher $k$.} \label{recall@k}
\end{figure}

Finally, we perform a ranked evaluation since the difficulty of the ranking task varies per patient due to 3 main factors:

 \textbf{Pool size}: Each patient exam results in a different number of WSIs (Fig.~\ref{recall@k}A). The top-$k$ slide selection task becomes more difficult as the number of slides increases. Therefore, slide triage from a larger pool gets a higher weight:
    \begin{equation}
        \texttt{pool\_reward}_i = \big( \log({\texttt{pool\_size}_i}) / \log(\texttt{MAX\_POOL\_SIZE}) \big)^{0.5}
    \end{equation}
    
\textbf{Annotated set size}: Coarse annotations available at the part level present an easier evaluation task. Conversely, fine-grained annotations consisting of one or a few block codes present a more difficult evaluation task. Therefore, smaller annotated sets are assigned a higher weight:
    \begin{equation}
        \texttt{set\_size\_reward}_i = \frac{\log({\texttt{pool\_size}_i - \texttt{set\_size}_i + 1})}{\log(\texttt{pool\_size}_i)}
    \end{equation}
The $+1$ term ensures a positive numerator and that \texttt{set\_size\_reward} is 0 when the \texttt{set\_size} equals the \texttt{pool\_size}, in which case any ranking would yield block codes in the annotated set.
    
\textbf{Rank}: Finally, higher rank for correctly triaged slide(s) is assigned higher weight: 
    \begin{equation}
        \texttt{rank\_penalty}_i = 1 / \log_2({\texttt{rank}_i + 1})
    \end{equation}
    The $+1$ in the denominator ensures that there is no division by $0$.
The final \texttt{score} for the $i$th patient is simply:
    \begin{equation}
        \texttt{score}_i = \texttt{pool\_reward}_i \times \texttt{set\_size\_reward}_i \times \texttt{rank\_penalty}_i
    \end{equation}

We ignore all patients with a single slide since $\texttt{pool\_size} = 1$ does not present a selection task. The \texttt{score} is bounded in the range $[0,1]$, with $1$ being the optimal score, and final scores are averaged across the dataset for each $k$.

\section{Results}

\subsection{All models achieve high recall}
All models show similar performance on the HN dataset with CONCH (top recall of $97.89\%$ at $k=8$) performing marginally better than mean-CONCH and TITAN (top recall of $97.74\%$ at $k=9$ and $97.31\%$ at $k=8$, respectively) (Fig.~\ref{recall@k}B). Since the HN dataset is coarsely labeled with annotated sets containing on average $15.2 \pm 10.1$ block codes, it presents an easier evaluation task. This is reflected in the high recall at $k=1$ ($\sim 92\%$). Evaluation on the HN-node subset, which contains more fine-grained labels for involved lymph nodes with annotated sets containing on average $3.6 \pm 3.4$ block codes, yielded a recall of $100\%$ at $k=1$ for all models. This performance on a more realistic evaluation task indicates that pathology FMs can accurately identify tumor content in slides. Visualizing patch-level classification generated by CONCH quantitatively demonstrates accurate identification of slide morphology, as confirmed by a pathologist (Fig.~\ref{recall@k}C). 

Performance on the OV dataset, which also presents a more realistic evaluation task with annotated sets containing an average of $1.9 \pm 1.1$ block codes, is worse at lower $k$. However, at higher $k$, performance of all models matches the performance observed on the easier HN dataset. When evaluated against primary blocks alone, recall tops out at $85\%$ at $k=7$ for CONCH and TITAN and at $k=9$ for mean-CONCH (Fig.~\ref{recall@k}D, left). When evaluated against primary and secondary blocks, we achieve recall of $100\%$ at $k=7$, $9$, and $6$ for CONCH, mean-CONCH, and TITAN, respectively (Fig.~\ref{recall@k}D, right). This result implies that including the top $9$ slides ranked by any of these models guarantees the inclusion of at least one slide containing tumor. Repeating the evaluation with the dataset filtered to remove parts identified in the pathology report to be negative for tumor yielded minimal changes, with TITAN achieving a recall of $100\%$ at $k=5$ when evaluated against primary and secondary blocks, demonstrating that these models accurately identify slides without tumor.

\subsection{CONCH and mean-CONCH show highest agreement}
\begin{figure}[t]
\includegraphics[width=0.95\textwidth]{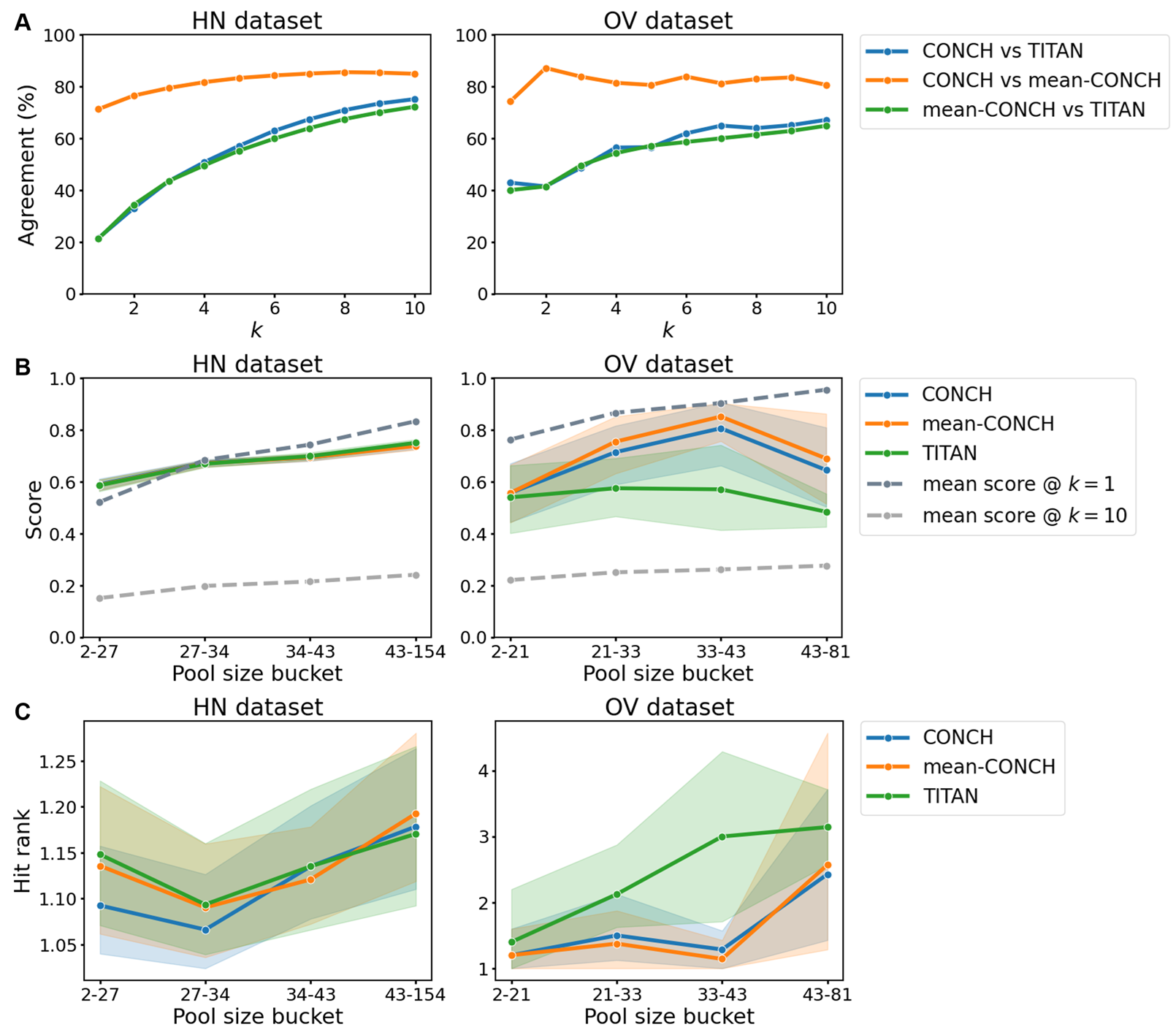}
\caption{A: Agreement between rankings from CONCH and mean-CONCH is highest on both the HN and OV datasets. B: Ranked evaluation shows comparable performance across models on the HN dataset, while CONCH and mean-CONCH show better performance on the OV dataset. C: Hit rank is also similar across models on the HN dataset, while CONCH and mean-CONCH show better performance on the OV dataset, with the tumor slide identified within the top-$2$ ranked slides for patients with up to $43$ slides.} \label{agreement_rankedeval}
\end{figure}

Agreement between CONCH and mean-CONCH is highest across both datasets, as is expected since they represent different aggregations of the same embeddings. Agreement between CONCH/mean-CONCH and TITAN is lower, implying that there is a mismatch between the top-$k$ slides ranked by TITAN when compared to those ranked by CONCH and mean-CONCH, especially at lower $k$ (Fig.~\ref{agreement_rankedeval}A). 

\subsection{Ranked evaluation score highest for CONCH and mean-CONCH}
Ranked evaluation shows each model's performance across pool sizes (number of slides per patient), which were bucketed into quartiles to ensure roughly equal-sized buckets. We also computed performance using the mean pool size and mean annotated set size per bucket at $k=1$ and $10$ to determine the expected performance for each dataset (Fig.~\ref{agreement_rankedeval}B). The expected performance at $k=1$ is lower for the HN dataset than for the OV dataset due to the larger annotated set sizes in the HN dataset. All models show similar performance on the HN dataset due to the coarsely labeled annotated set, which induces a higher likelihood that block codes from the annotated set appear at or near $k=1$. In the smallest pool-size bucket, all models perform better than the expected performance at $k=1$. On the OV dataset, CONCH and mean-CONCH outperform TITAN, with all models showing a drop in performance at the largest pool sizes. The hit rank, or $k$ at which a block code from the annotated set is identified is, on average, $1.4$ and $1.3$ across the $3$ smaller pool-size buckets for CONCH and mean-CONCH, respectively (Fig.~\ref{agreement_rankedeval}C). However, the hit rank drops to $2.4$ and $2.6$ in the largest bucket for CONCH and mean-CONCH, respectively. Average hit rank for TITAN, on the other hand, drops consistently from $1.5$ to $2$, $3$ and $3.1$ as pool sizes increase (Fig.~\ref{agreement_rankedeval}C).

\section{Discussion}
Automatic triage of digital pathology slides is critical not only for streamlining pathologist evaluation of cancer patient cases, but also for curating and labeling datasets for model training. In this paper, we show that publicly available FMs trained jointly on pathology image and text data can successfully triage slides without manual curation. Patch-level FMs like CONCH show better performance than slide-level FMs like TITAN, which may be because patch-level FMs classify slides at a more granular level and therefore may be better able to identify smaller tumors or tumors in slides where lymphocytes or other cell types are more dominant. Slide-level FMs, like TITAN, on the other hand, must aggregate these fine-grained patch embeddings into a single slide embedding, potentially suppressing some features present in the slide. This may also explain the disagreement between TITAN and CONCH/mean-CONCH, as well as the lower average hit rank achieved by TITAN for patients with larger sets of slides. 

Future work includes evaluation of this pipeline using additional models and exploration into differences in performance between the studied models. For instance, exploration into representations learned by different models using mechanistic interpretability methods could shed light on the differences in performance. Further, evaluation may be performed on additional datasets to study generalizability across cancer types. Preliminary results on survival prediction using the top-$1$ slide ranked by CONCH show improvement over predictions using all slides per patient~\cite{yao2020whole,shao2023hvtsurv}, and evaluation using top-$k, k>1$ slides is ongoing.

\begin{credits}
\subsubsection{\ackname} This study was funded by the Mayo Clinic Comprehensive Cancer Center, Susan Morrow Legacy Foundation, and Heidi Diercks Krause Fund in AI Innovation for Cancer. We would also like to thank Jake Matras and Dr. Derek Alexander for their feedback, and Suzanne Lee-Trejo for managing collaboration across teams.

\subsubsection{\discintname}
The authors have no competing interests to declare that are relevant to the content of this article.
\end{credits}

%
%
%
\bibliographystyle{splncs04}
\bibliography{refs}

\section{Appendix}
\setcounter{table}{0}
\renewcommand{\thetable}{A\arabic{table}}

The appendix contains details about the prompts used for zero-shot classification.

\begin{table}[h]
\centering
\caption{Prompts used for \texttt{tumor} class for the HN and OV datasets.}
\begin{tabularx}{\textwidth}{p{1cm}|C}
\hline
 & \textbf{\texttt{tumor}}  \\
\hline
\textbf{ HN} & "squamous cell carcinoma",
                "head and neck squamous cell carcinoma",
                "HNSCC",
                "head and neck SCC",
                "oropharyngeal squamous cell carcinoma",
                "laryngeal squamous cell carcinoma",
                "hypopharyngeal squamous cell carcinoma",
                "nasopharyngeal squamous cell carcinoma",
                "oral cavity squamous cell carcinoma",
                "HNSC" \\
\hline
\textbf{ OV} & "high-grade serous ovarian cancer",
                "high-grade serous carcinoma",
                "HGSOC",
                "serous papillary carcinoma",
                "serous ovarian carcinoma" \\
\hline 
\end{tabularx} \label{tumor_prompts}
\end{table}

\begin{table}[h]
\centering
\caption{Prompts used for \texttt{lymphocyte} class for the HN and OV datasets.}
\begin{tabularx}{\textwidth}{p{1cm}|C}
\hline
 & \textbf{\texttt{lymphocyte}} \\
\hline
\textbf{ HN} & \multirow{4}{\hsize}{\centering "lymphocytes",
                "lymphoid aggregate",
                "immune cells",
                "lymphoid infiltrate",
                "inflammatory cells",
                "tumor-infiltrating lymphocytes",
                "TILs",
                "tumor-infiltrating immune cells"} \\
& \\
\cline{1-1}
\textbf{ OV} & \\
& \\
\hline
\end{tabularx} \label{lymphocyte_prompts}
\end{table}

\begin{table}[h]
\centering
\caption{Prompts used for \texttt{other} class for the HN and OV datasets.}
\begin{tabularx}{\textwidth}{p{1cm}|C}
\hline
 & \textbf{\texttt{other}}  \\
\hline
\textbf{ HN} & "resection negative for carcinoma",
                "resection negative for tumor",
                "normal oropharyngeal mucosa",
                "uninvolved oropharyngeal mucosa",
                "normal oropharyngeal epithelium",
                "benign epithelium" \\
\hline
\textbf{ OV} & "resection negative for carcinoma",
                "resection negative for tumor",
                "normal ovarian tissue",
                "uninvolved ovarian tissue",
                "normal ovarian epithelium",
                "benign epithelium" \\
\hline
\end{tabularx} \label{other_prompts}
\end{table}

\begin{table}[t]
\centering
\caption{Templates used to generate an ensemble of prompts. Prompts were generated by combining template text with class prompts by replacing `CLASSNAME' with the class prompts shown in Tables~\ref{tumor_prompts},~\ref{lymphocyte_prompts},~\ref{other_prompts}. Adapted from example in the CONCH repository.}
\begin{tabularx}{\textwidth}{C}
\hline
\\
"CLASSNAME.",
            "a photomicrograph showing CLASSNAME.",
            "a photomicrograph of CLASSNAME.",
            "an image of CLASSNAME.",
            "an image showing CLASSNAME.",
            "an example of CLASSNAME.",
            "CLASSNAME is shown.",
            "this is CLASSNAME.",
            "there is CLASSNAME.",
            "a histopathological image showing CLASSNAME.",
            "a histopathological image of CLASSNAME.",
            "a histopathological photograph of CLASSNAME.",
            "a histopathological photograph showing CLASSNAME.",
            "shows CLASSNAME.",
            "presence of CLASSNAME.",
            "CLASSNAME is present.",
            "an H\&E stained image of CLASSNAME.",
            "an H\&E stained image showing CLASSNAME.",
            "an H\&E image showing CLASSNAME.",
            "an H\&E image of CLASSNAME.",
            "CLASSNAME, H\&E stain.",
            "CLASSNAME, H\&E." \\
\\
\hline
\end{tabularx} \label{prompt_templates}
\end{table}

\end{document}